\documentclass[conference]{IEEEtran}
\IEEEoverridecommandlockouts
\usepackage{cite}
\usepackage{amsmath,amssymb,amsfonts}
\usepackage{algorithmic}
\usepackage{graphicx}
\usepackage{textcomp}
\usepackage{xcolor}
\usepackage{stfloats}
\usepackage{multirow}
\usepackage{subfigure}
\def\BibTeX{{\rm B\kern-.05em{\sc i\kern-.025em b}\kern-.08em
    T\kern-.1667em\lower.7ex\hbox{E}\kern-.125emX}}

\DeclareRobustCommand*{\IEEEauthorrefmark}[1]{%
        \raisebox{0pt}[0pt][0pt]{\textsuperscript{\footnotesize\ensuremath{#1}}}}
\begin{document}

\title{Crowd Video Captioning\\
\thanks{This work is supported in part by the National Science Foundation China (NSFC)
61761136005, and ARC through DP190100887 and DP160104500.}
}


\author{
\IEEEauthorblockA{Liqi Yan\IEEEauthorrefmark{1,2}, Mingjian Zhu\IEEEauthorrefmark{1,3}, Changbin Yu\IEEEauthorrefmark{*}\IEEEauthorrefmark{1}}
\IEEEauthorblockA{\IEEEauthorrefmark{*}Corresponding email: yu\_lab@westlake.edu.cn} 
\IEEEauthorblockA{\IEEEauthorrefmark{1}L. Yan, M. Zhu and C. Yu are with the School of Engineering, Westlake University, China} 
\IEEEauthorblockA{\IEEEauthorrefmark{2}L. Yan is also enrolled at Fudan University, China} 
\IEEEauthorblockA{\IEEEauthorrefmark{3}M. Zhu is also enrolled at Zhejiang University, China} 
}

\maketitle

\begin{abstract}
	  Describing a video automatically with natural language is a challenging task in the area of computer vision. In most cases, the on-site situation of great events is reported in news, but the situation of the off-site spectators in the entrance and exit is neglected which also arouses people's interest. Since the deployment of reporters in the entrance and exit costs lots of manpower, how to automatically describe the behavior of a crowd of off-site spectators is significant and remains a problem.
	  To tackle this problem, we propose a new task called crowd video captioning (CVC) which aims to describe the crowd of spectators. We also provide baseline methods for this task and evaluate them on the dataset WorldExpo'10.  Our experimental results show that captioning models have a fairly deep understanding of the crowd in video and perform satisfactorily in the CVC task. \\

\end{abstract}


\section{Introduction}
With the rapid development of the deep neural network, computers can describe the content of the video in a reasonably deep way. Video captioning has important practicality and wide potential application, and a typical one is news broadcasting. 

In most great events, there are professional commentators in the stadium to broadcast the situation in real time, and some studies about automatic sports video commentary \cite{b3,b4} have been carried out. Outside the venue, entry and exit of spectators are also important. Reports such as \textit{``The line of people snaked into the theater with joy in their faces."} often appear in the news, but deliberately assigning reporters to wait for spectators wastes manpower. Therefore, to report the situation of off-site spectators in real time, we use the surveillance camera to analyze the spectators' crowd and generate captions with the deep learning methods.

Recently, there has been some works on crowd counting \cite{b5,b6,b7} and classification \cite{b8,b9,b10,b11}, but their output is a number of pedestrians, or the state of mobility and abnormal behaviors. None of these work can produce a descriptive sentence for the crowd as a news report. 

A caption of crowd video needs to describe various attributes of a crowd, such as the number of people in the crowd, the situation of movement, direction of flow, etc. Therefore, we use a captioning framework to describe the crowd in videos. In major events, if the surveillance camera uses our system, it can be directly connected to the news broadcasting system to broadcast the real-time off-site situation to the news media.

Our framework uses convolutional neural networks to extract these crowd features, then feeds them into a classifier or a language model to produce the summary. All attributes and situations of the crowd should be included in the output descriptive words of the language model.

To validate our system, we create a crowd video captioning dataset, which is based on the crowd counting dataset: WorldExpo'10. We select some of the videos in this dataset and make captions for them. Several experiments using the proposed models have been carried out to evaluate the performances of those methods. 

The main contribution of our work is the proposal of a new task called crowd video captioning \textbf{(CVC)} which aims to generate captions for the crowd video. We provide baselines and a system framework for this task, and the results of the experiments prove the feasibility of our system.

\section{Related Work}

\subsection{Crowd Counting}

In recent years, many models and datasets have been proposed for crowd counting. For example, Chan et al.~\cite{b5} collect UCSD in the University of California, San Diego, and it is one of the earliest datasets for crowd counting. Chan et al.~\cite{b5} have used Dynamic Textures and Gaussian methods to count the crowd in videos. After that, many new models have been proposed for this task, such as CNN \cite{b6} and ACSVP~\cite{b7} (a GAN-based, U-net structured model).

WorldExpo'10 proposed in \cite{b6} is another large-scaled dataset for crowd counting. It includes more than a thousand labeled videos captured by over one hundred monitoring cameras, all from the Shanghai World Expo in 2010. We have used it in this research.

\subsection{Crowd Behaviors Analysis}

In addition to counting, researches on behavior analysis of crowd are also underway. MED \cite{b8} has been carried out as the crowd emotion dataset, but it only has 31 videos, and the people in the crowd are just walking around and making some specific movements, such as fighting, hugging. 

The newest dataset, Crowd-11 proposed in \cite{b10}, has been provided to classify the fine-grained crowd behaviors. It categorizes the flow mainly by the direction of each one in the crowd. Models including LSTM \cite{b9}, C3D, V3G \cite{b10}, ConvLSTM \cite{b11} and so on, have been used to analyze the crowd abnormal behaviors in those datasets. While these systems work efficiently, they do have significant disadvantages: the accuracy of fine-grained classification for flow is generally low and they are more suitable for abnormal behavior monitoring.

\subsection{Fine-grained Video Captioning}

There are also some works about fine-grained video captioning, including broadcasting for tennis videos \cite{b3}, and Fine-grained Sports Narrative dataset \cite{b4}. 
Models like LSTM-YT \cite{b12} and S2VT \cite{b13} have been used to complete those tasks. But the aims of those works are all for professional sports broadcasting.

\section{Methods}

A conventional video captioning pipeline can be divided into two stages: feature extraction and caption generation. 

In the first stage, we can use the model for image classification as a frame feature extractor, or the model for video classification as a video feature extractor. The features extracted by these two methods are different, so suitable models should be selected for different tasks..

In the second stage, a sequence to sequence model is fed with the extracted features and generate sentences. Therefore, a language model such as a recurrent neural network (RNN) can be used to construct this decoder.

\subsection{Frame Feature Extraction}


The high-level 2D features of the frame can be extracted by a convolutional neural network (CNN) for image classification, feeding it with frames of the video one after another, and getting the feature of each frame from the last layer of the model.

Inception V3 \cite{b14} is one of those networks, evolved from GoogLeNet \cite{b15}. It decomposes a 2D convolution mask which is $n\times n$ into two 1D masks which are $1\times n$ and $n\times 1$ respectively. It can not only accelerate the calculation but also increase the depth of the network.

In addition to Inception V3, ResNet \cite{b16} and Inception V4 \cite{b14} are also available for the frame feature extraction. Their inputs have a direct influence to the output through skip connections.

\subsection{Video Feature Extraction}

C3D \cite{b17} is a network for extracting features from videos. Unlike the 2D convolution in 3D space which can not slide in the temporal domain, 3D convolution can extract the features of the same region in different periods.

\begin{figure}[htbp]
	\centerline{\includegraphics[width=3.5in]{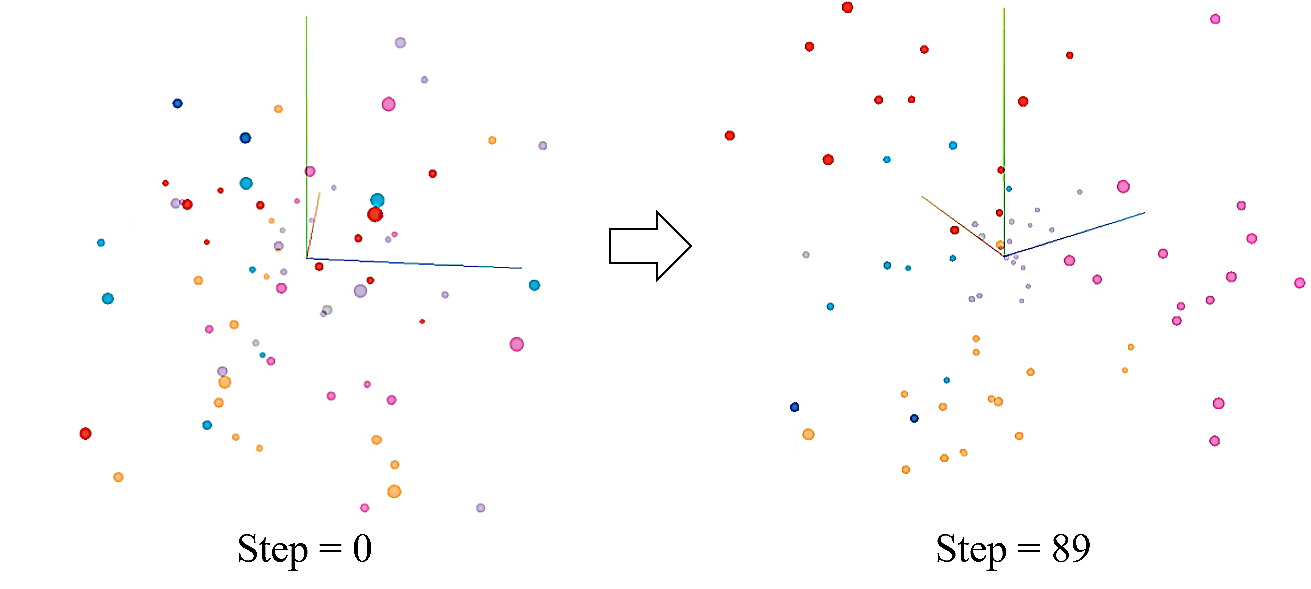}}
	\caption{The output of C3D model. (Best View in color)}
	\label{fig1}
\end{figure}


C3D network has eight convolution and five pooling layers. Using Principal Component Analysis (PCA), the output of the last layer (fc7) in C3D is shown in Fig.~\ref{fig1}, where every point represents the feature of its corresponding video, and the 8 different colors represent 8 different classes. Before training (step=0), all points from different categories are mingled together. But after training (step=89), points of the same category are collected together. There are 70 points in the figure, the length of their tensor is 4096.

\subsection{Caption Generation}

\begin{figure*}[htbp]
	\centerline{\includegraphics[width=6.2in]{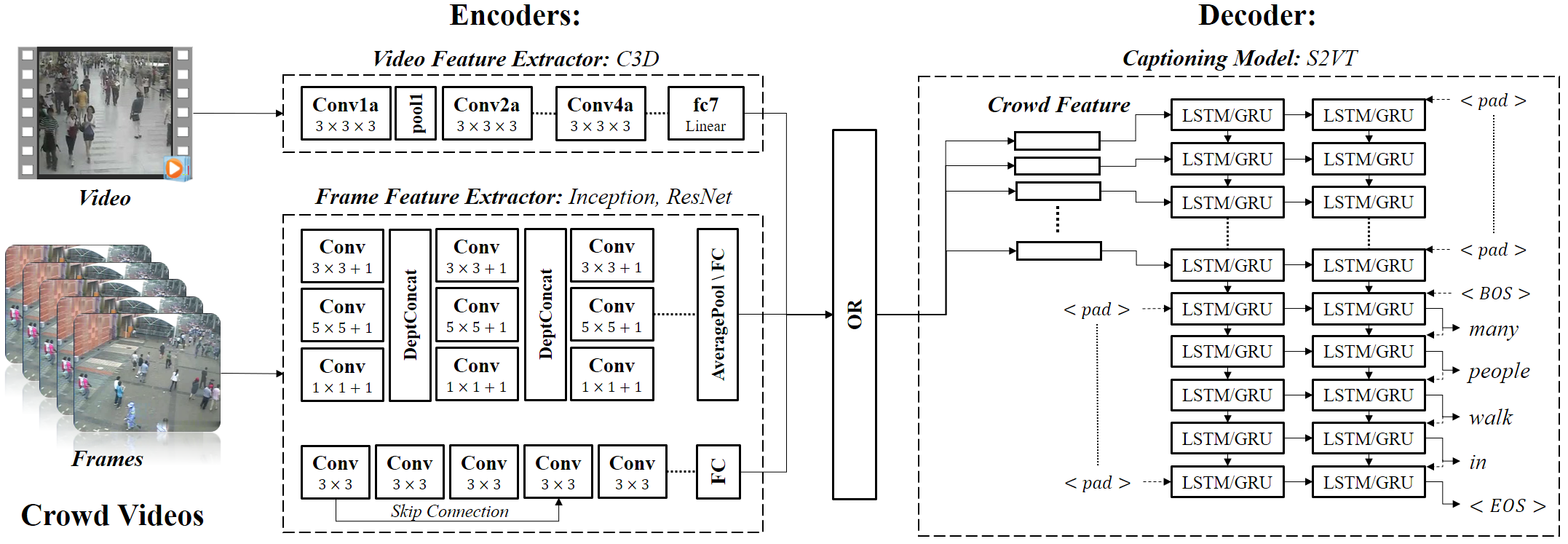}}
	\caption{Our proposed framework for \textbf{CVC}}.
	\label{fig3}
\end{figure*}

S2VT \cite{b13} is a typical sequence-to-sequence network for generating captions for videos. It's made up of two-layer recurrent neural network (RNN), and long short term memory (LSTM) \cite{b18} is used as the cells of this RNN.

As Fig.~\ref{fig3} shows, S2VT takes the features extracted from the video frames as the input sequences. The ``words" of those embeddings are fed into the LSTM cells of the first layer one by one, then, after several iterations, the words for captioning are continuously generated by the second layer.

S2VT generates words by taking the words which are already generated. It uses $\textless BOS\textgreater$ to indicate the begin-of-sentence and $\textless EOS\textgreater$ for the end-of-sentence tag. And $\textless pad\textgreater$ is used when there is no input at the time step. Those labels are all utilized as the references to produce the next word.

\section{Details of Models}

In this section, we first introduce the definition of our crowd video captioning task and the overview of our frameworks. Then, we describe two alternative models to caption the crowd video.

\subsection{Task Definition and Overview}

Crowd video captioning aims to describe $q$ attributes of a crowd in natural language, such as the number of people in the crowd, the situation of movement, the direction of flow, etc. Our model can be divided into two parts: encoder and decoder. First, we need to identify the crowd from videos. It is easy to detect the changing areas between frames because the pixels representing pedestrians tend to move together as a whole. Then, $n$ frames are randomly selected from a video. the attributes and situations of the crowd can be extracted from the following features: crowd extent size, density, individual movements, pedestrian situations and so on. Our framework employs convolutional neural networks as the feature extractor to get these features embeddings.

The features extracted from the videos include the attributes of the crowd. Therefore, crowd's feature can be directly extracted from the extractor frame by frame, and then joined into one sequence $F_{frame}=(f_1,f_2\cdot \cdot \cdot f_n )$. Crowd's features can also be encoded into a vector sequence $F_{video}=(x_1, x_2\cdot \cdot \cdot x_t)$ through the video feature extractor. Finally, this sequence is fed to a classifier or a language model as the caption generator to produce the description $Y=(y_1, y_2, \cdot\cdot\cdot y_m)$ of the crowd.

\subsection{Classification Model}

If there are $t$ words in the dataset, the number of $m$-word sentences can be formed theoretically is $t^m$. But very few of those randomly generated sentences are grammatical. So we use $p$ grammatical sentences as the tags that classifier needs to recognize. In this model, after the video features $\boldsymbol{F}_{video}=(x_1, x_2\cdot \cdot \cdot x_t)$ are extracted from the encode, a classifier is used as the decoder. 

\begin{figure}[htbp]
	\centerline{\includegraphics[width=3.1in]{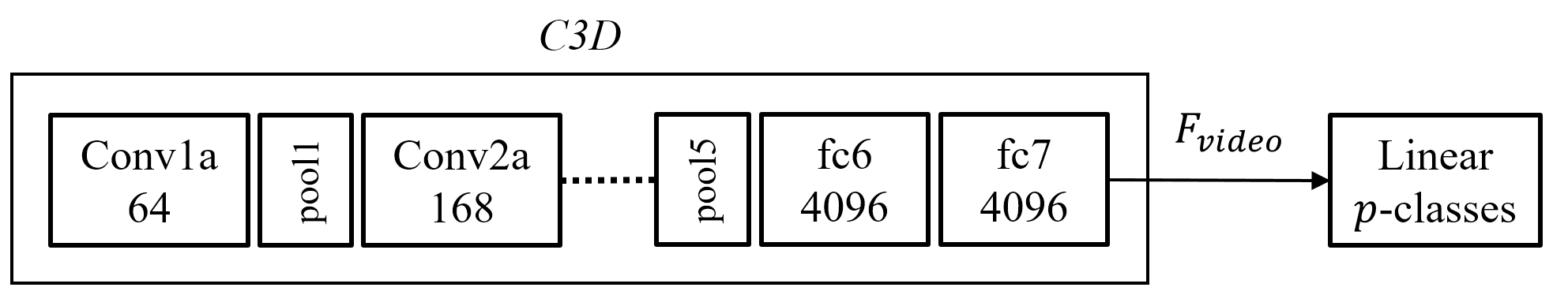}}
	\caption{The C3D model and the linear layer following it.}
	\label{fig2}
\end{figure}

Let's take p-category classifier based on C3D as example, a linear layer is used as the classifier followed in the rear of C3D, as shown in Fig.~\ref{fig2}. The linear layer converts $t=4096$ dimensional vectors $\boldsymbol{F}_{video}=(x_1,x_2,\cdot\cdot\cdot ,x_{4096})$ to p-dimensional outputs, then the probabilities $\boldsymbol{Z}=(z_1,z_2,\cdot\cdot\cdot ,z_p)$ for which label to be output can be calculated by \emph{softmax} as follow:
\begin{equation}
z_j=softmax(x_j)=\frac{e^{x_j}}{\sum_{j=1}^pe^{x_j}}
\end{equation}
where $j\in{\{x|1\leq x\leq p, x\in\mathbb{N}\}}$, $p$ is the total number of categories. 
After that, the corresponding value of the category with the maximal probability is set to 1.

\subsection{Captioning Model}

Since classification model can only generate captions within the label set, we adopt the captioning model in our \textbf{CVC} system framework. To caption the video, our captioning model needs to generate natural language sequences. Given an frame feature sequence $\boldsymbol{F}_{frame}=(f_1, f_2\cdot \cdot \cdot f_n)$, the goal of captioning model is to generate proper words: $\boldsymbol{Y}=(y_1,y_2\cdot\cdot\cdot y_m)$. The model estimates the probability:
\begin{equation}
P(Y|\boldsymbol{F}_{frame})=\prod_{i=1}^{m}P(y_i|y_{<i},\boldsymbol{F}_{frame})
\end{equation}
where $y_{<i}$ represents all the generated words before the $i^{th}$ word. When captioning model outputs each word, it chooses the word with the highest probability of this position on the basis of all the words that have been output before.

The overall design of our \textbf{CVC} system framework of captioning model is shown in Fig.~\ref{fig3}. 

Firstly, the video frames feature $\boldsymbol{F}_{frame}$ or the whole video feature $\boldsymbol{F}_{video}$ about the crowd is obtained from the encoder, such as C3D, ResNet or Inception. 

After that, we take the language model, like S2VT shown in Fig.~\ref{fig3}, as the decoder for captioning. In addition to the LSTM, we apply a gated recurrent neural network (GRU) \cite{b19} as the cell of S2VT, which has fewer parameters than LSTM and can avoid over-fit. Researches and practical experiences show that these two cells each have their own advantages and disadvantages.

The following steps show how the model produces the captions. 

First of all, vocabulary which contains all the words in the dataset is built, and every word is encoded into a vector. 

Next, the features $\boldsymbol{F}_{frame}$ are input to the model, and then which word $y_i$ to be selected from the vocabulary depends on the hidden state $h_{n+i-1}$ and the network parameters $\theta$:

\begin{equation}
y_i^*=\mathop{\arg\max}_{y_i} \sigma(y_i,h_{n+i-1},\theta)
\end{equation}

Finally, supposing that all the hidden states of the network are $\boldsymbol{H}=(h_1, h_2\cdot\cdot\cdot h_{n+m})$, from the beginning of the input sentence to the end of the output sentence, the model optimizes the parameters by maximizing the sum of a log-likelihood probability of the generated words as follow:
\begin{equation}
\theta^*=\mathop{\arg\max}_{\theta} \sum_{i=1}^{m}\log P(y_i|h_{n+i-1},y_{i-1};\theta)
\end{equation}

\subsection{Loss Function}

During training, we use the Cross-Entropy Loss as the loss function, which is defined as follow:
\begin{equation}
\boldsymbol{L}=-[\boldsymbol{Z}\log \boldsymbol{\hat{Z}}+(1-\boldsymbol{Z})\log(1-\boldsymbol{\hat{Z}})]
\end{equation}
where $\boldsymbol{Z}$ is the ideal embedding of the words created by humans and the $\boldsymbol{\hat{Z}}$ is the computer-generated words embedding. The greater the difference between the predicted results and the ground truth is, the higher the gradient of the loss function is, and the faster the convergence rate is.

\section{Experiments}

To caption the crowd videos, we select WorldExpo'10 from a series of datasets for crowd analysis. And in order to evaluate our baseline and captioning methods in our dataset, C3D and other feature extractors have been chosen as the encoder, and experiments on S2VT with LSTM and GRU have been token.

\subsection{Dataset}

Crowds in most videos of the WorldExpo’10 dataset are messy because this dataset is mainly used for crowd counting. In order to simplify the task, we select 98 videos of it and caption them based on the crowd. These videos are captured by 7 surveillance cameras. 

\begin{figure}[htbp]
	\centerline{\includegraphics[width=3in]{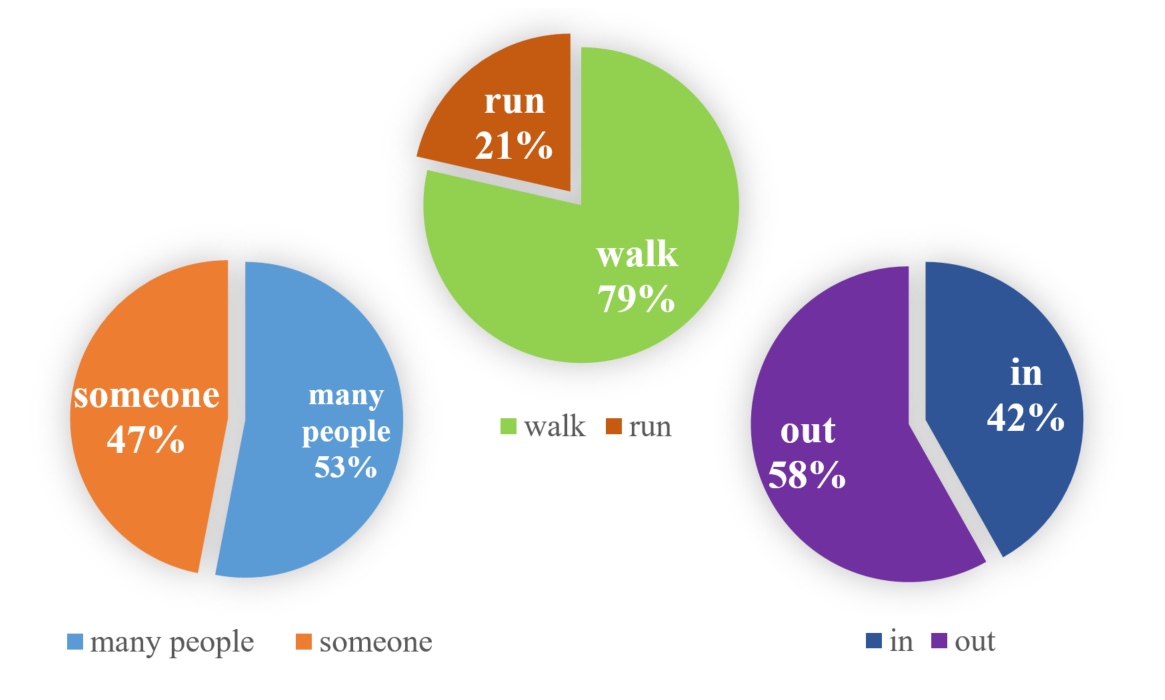}}
	\caption{The composition of my dataset. (Best View in color)}
	\label{fig4}
\end{figure}

Keywords of captions in our dataset are shown in Fig.~\ref{fig4}, they describe the number of people in the crowd, the situation of movement and the direction of flow respectively. Because ``running" accounts for a small percentage in the WorldExpo'10 dataset, so it only accounts for 21\% in our dataset.  We define the direction to be close to the camera as ``in", and the direction to be far away from the camera as ``out".

The size of the vocabulary is 6, the number $q$ of attribute pairs that make up the descriptive sentences is 3. So $p=2^q=8$ captions are formed in our dataset.

\subsection{Baseline Based on C3D}

The first baseline is using C3D directly as $p$-category classifier. Since the words in our dataset can make up $p=2^q=8$ sentences, we divide all the videos into eight categories. The label of each category is one of these sentences, such as ``Many people walk in". 

The dimension of the last layer in C3D is $t=4096$, we add a linear full connected layer after it as the classifier. The dimension of the output of this linear layer the total number of categories, which is 8 in our experiment.

The model is first pre-trained with UCF101 \cite{b20}. We then fine-tune the network on our dataset. The split for training, validating and testing is 70:19:9. In order to fit the C3D model, frames are resized into $128\times171$, and the randomly cropped to $112\times112$.

\begin{figure*}[htbp]
	\centering
	\subfigure[\textbf{Accuracy} curve for \textbf{training} epochs]
	{\includegraphics[width=1.8in]{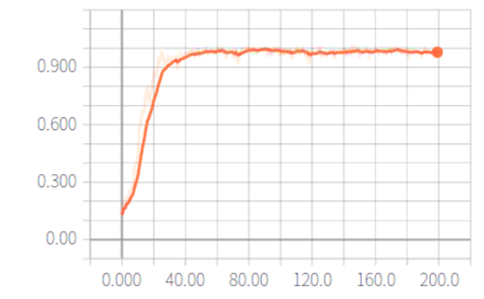}}
	\subfigure[\textbf{Accuracy} curve for \textbf{validating} epochs]
	{\includegraphics[width=1.8in]{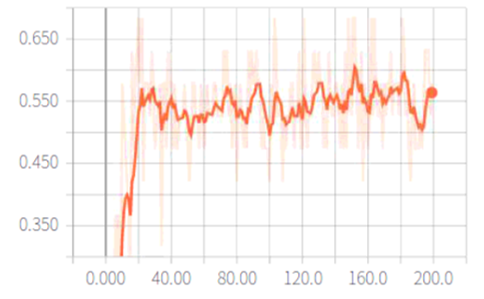}}
	\subfigure[\textbf{Accuracy} curve for \textbf{testing} epochs]
	{\includegraphics[width=1.8in]{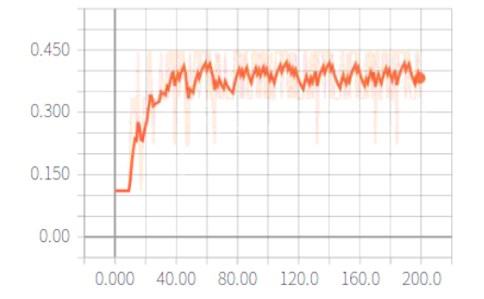}}
	\subfigure[\textbf{Loss} curve for \textbf{training} epochs]
	{\includegraphics[width=1.8in]{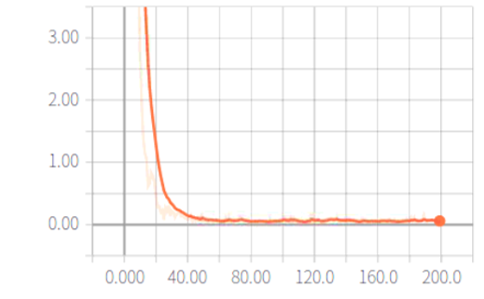}}
	\subfigure[\textbf{Loss} curve for \textbf{validating} epochs]
	{\includegraphics[width=1.8in]{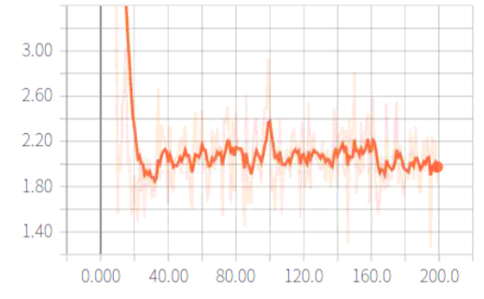}}
	\subfigure[\textbf{Loss} curve for \textbf{testing} epochs]
	{\includegraphics[width=1.8in]{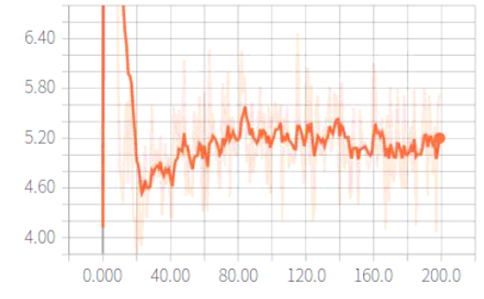}}
	\caption{Accuracy and loss curves for different epochs. (Best View in color)}
	\label{fig5}
\end{figure*}

The accuracy and loss curves for training, validating and testing epochs are shown in Fig.~\ref{fig5}, where the number of frames inputted to the model is 16, the learning rate is set to $10^{-4}$, and the schedule is set to divide the learning rate by 2 every 10 epochs. And the curves in Fig.~\ref{fig5} are smoothed by 0.8, while the original values are reported in faint polylines.

The curves show that in the training epochs, the loss almost converges to zero, and the accuracy can achieve fast convergence as well. It reaches 0.9714 and 0.6842 on training and validating corpus, but it's only 0.4444 on testing one. Their loss can be reduced to approximately zero on the training set, but not on the validation set or the test set.


\subsection{Evaluation Metrics for Captioning}

In this section, we introduce several frequently used evaluation metrics for video captioning.
\begin{itemize}
	\item \textbf{BLEU.} It is based on modified n-gram precision. To begin with, the modified precision $p_n$ is defined as the candidate counts clipped by their corresponding reference maximum value, summed, and divided by the total number of candidate n-grams. In the second place, supposing $c$ is the length of the candidate translation and $r$ is the effective reference corpus length, we compute the brevity penalty:
	\begin{equation}
		BP=\left\{
			\begin{array}{lcl}
				1, & c > r. \\
				e^{(1-\frac{r}{c})}, &  c\leq r.
			\end{array}
		\right.
	\end{equation}
	Ultimately, we use n-grams from 1-gram up to length N to calculate the BLEU score with the weighted precision:
	\begin{equation}
		BLEU=BP\cdot \exp(\sum_{n=1}^{N}w_n\log(p_n))
	\end{equation}

	\item \textbf{CIDEr.} It measures the similarity of a sentence to the majority, or consensus of how most people describe the image. For instance, sentences such as ``Mike has a baseball and Jenny has basketball" is more representative of the consensus descriptions than the sentence ``Jenny brought a bigger ball than Mike". CIDEr is proposed to capture those sentences with more broad consensus.
	\item \textbf{METEOR.} It is based on the weighted precision $P$ and recall $R$ of the matched content-function words in hypothesis and reference. 
	 Fragmentation penalty $Pen=\gamma\cdot(\frac{ch}{m})^\beta$ is defined to account for differences in word order, where the chunks $ch$ defined as a series of matches that is contiguous and identically ordered in both sentences, and $m$ is the average number of matched words over hypothesis and reference. 
	 After the parameterized harmonic mean $E_{mean}$ of $P$ and $R$ is calculated, the METEOR score is computed as follow:
	\begin{equation}
		METEOR=(1-Pen)\cdot E_{mean}
	\end{equation}
	It's always small than 1 even if the sentences predicted are the same as the references, because the $Pen$ never equals zero.
	
	\item \textbf{ROUGE.} It counts the number of overlapping units such as n-gram, word sequences, and word pairs between the predicted captions and the referential summaries.
\end{itemize}

\subsection{Details and Results of Model based on S2VT}

We use C3D, ResNet-152 and Inception V3 (V4) pre-trained on UCF101 \cite{b20} or ImageNet \cite{b21} as the feature extractor of every frame from our dataset, then train the S2VT model directly with those features. LSTM and GRU are used as the RNN cell of the S2VT. For the best performance, the split for training and testing is adjusted to 45:4. We follow the experimental setting from the default. We set the dimension of features of video frames to 2048 and set the number of hidden layers to 512.

\begin{table*}[htbp]
	\caption{Performances of S2VT model with different RNN cell and feature}
	\begin{center}
		\begin{tabular}{|c|c|c|c|c|c|c|c|c|}
			\hline
			\textbf{Metric}&\multicolumn{4}{|c|}{\textbf{BLEU(\%)}} & 
			\multirow{2}{*}{\textbf{CIDEr(\%)}} & \multirow{2}{*}{\textbf{METEOR(\%)}} &
			\multirow{2}{*}{\textbf{ROUGE\_L(\%)}} & \multirow{2}{*}{\textbf{Accuracy}} \\
			\cline{2-5} 
			\textbf{(n-gram)} & \textbf{\textit{1-gram}} & \textbf{\textit{2-gram}} & \textbf{\textit{3-gram}} & \textbf{\textit{4-gram}} & & & & \\
			\hline
			C3D and LSTM & 82.76 & 76.89 & 74.24 & 75.64 & 63.46 & 50.30 & 82.83 & 0.625 \\
			C3D and GRU & \underline{92.86} & \underline{88.84} & \underline{86.97} & \underline{90.06} & \underline{72.53} & 56.97 & 91.67 & \underline{0.75} \\
			ResNet152 and LSTM & 78.52 & 69.13 & 59.37 & 50.92 & 47.02 & 43.24 & 80.57 & 0.625 \\
			ResNet152 and GRU  & \underline{92.86} & \underline{88.84} & 83.97 & 81.63 & 70.44 & 56.97 & \underline{92.71} & \underline{0.75} \\
			Incept.v3 and LSTM & 86.21 & 83.54 & 81.27 & 80.95 & 72.07 & \underline{58.38} & 86.99 & \underline{0.75} \\
			Incept.v3 and GRU  & \textbf{96.43} & \textbf{95.71} & \textbf{94.34} & \textbf{95.73} & \textbf{81.15} & \textbf{67.81} & \textbf{95.83} & \textbf{0.875} \\
			Incept.v4 and LSTM & 82.09 & 81.62 & 80.68 & 84.35 & 68.65 & 57.90 & 83.33 & \underline{0.75} \\
			Incept.v4 and GRU  & \underline{92.86} & \underline{88.84} & 83.97 & 81.63 & 70.44 & 56.97 & \underline{92.71} & \underline{0.75} \\
			\hline
		\end{tabular}
		\label{tab1}
	\end{center}
\end{table*}

We report BLEU, CIEDEr, METEOR, and ROUGE\_L captioning scores for this method, the results on the testing set are provided in TABLE.~\ref{tab1}. The learning rate is set to $4\times 10^{-5}$, the scheduler is set to decay the learning rate by 0.8 every 200 epochs. 

Results obtained from the S2VT with different RNN cell and feature are shown in TABLE.~\ref{tab1}, best performances are presented in bold, and second best performances are underlined. For our task, the most appropriate feature extractor is inception V3, followed by C3D. And GRU has fewer parameters than LSTM, so it converges more easily than LSTM when the dataset size is small. It is the reason why GRU works better than LSTM on our dataset.

\begin{figure}[htbp]
	\centerline{\includegraphics[width=2.5in]{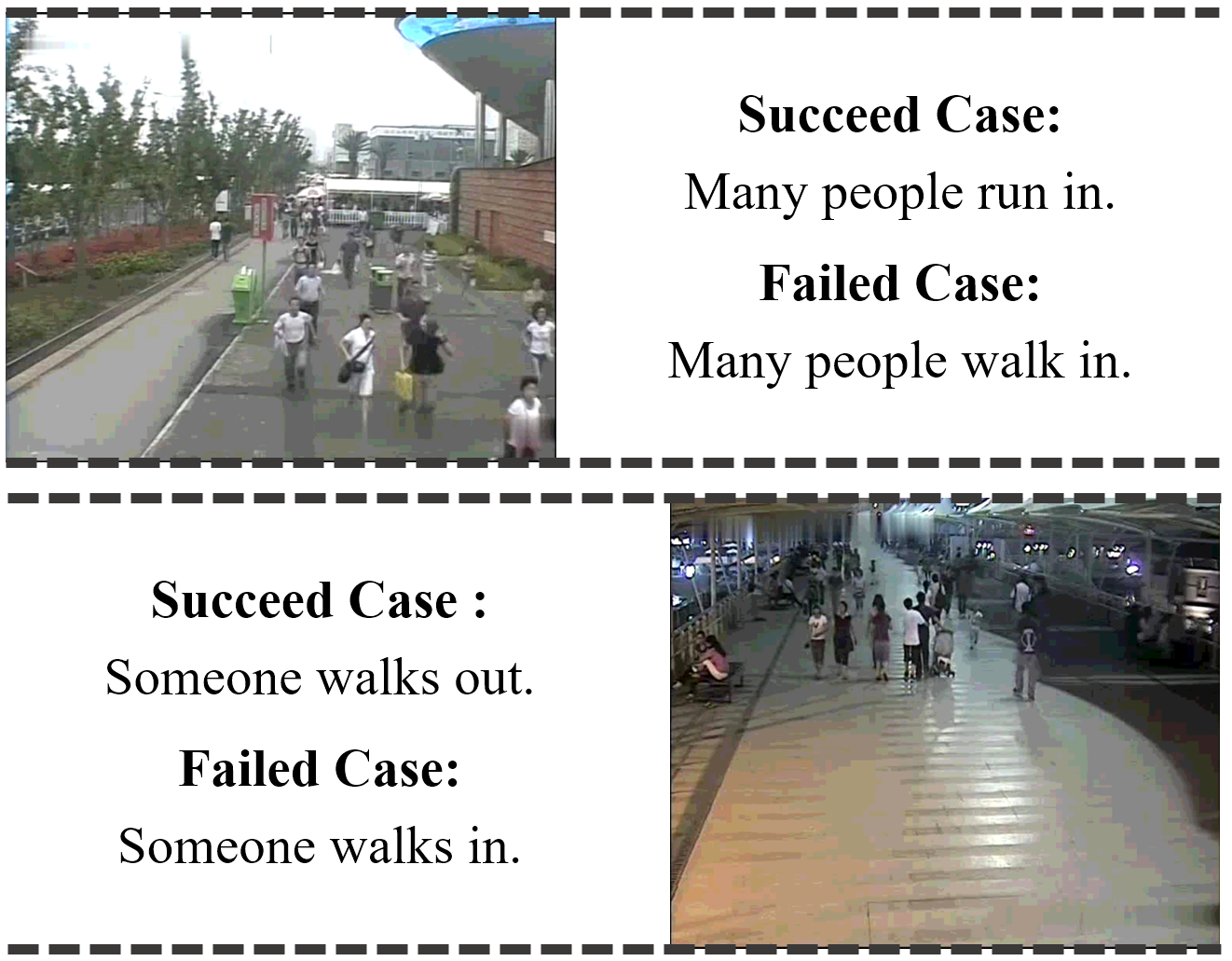}}
	\caption{The correct and incorrect results generated by captioning models.}
	\label{fig6}
\end{figure}

The accuracy is calculated in the purpose of being compared with the results of the C3D method, defined as the proportion of complete correct sentences to all of them. It is higher than the counterpart of the C3D classifier. And the two incorrect hypotheses generated by LSTM from Inception V3 features are just predicted with the mistake on verb and direction respectively, as shown in Fig.~\ref{fig6}. The second failed case is due to several people who interfere with the judgment. Although the sentence structure is not set manually, the output is completely consistent with the grammar, such as the singular-plural rule. 

\subsection{Experimental Conclusion}

Compared with the baseline, this method based on S2VT outperforms the classifier with C3D due to the comprehension of words in the sentence separately. This verifies that it is not necessary to know what the specific features represent when extracting features, and the analysis of crowd features can be directly handed over to the language model. 

Moreover, it proves that image classification models such as Inception and ResNet can extract the feature of the crowd in each frame, and the sequence composed of these features can be used for the crowd captioning.

This reflects the power of the video captioning model, especially if more information needs to be described in the summary later. Video captioning models can interpret the temporal information from the features extracted by the convolution neural network. It can also understand which information each word represents, and make them into a sentence without grammatical mistakes.

\section{Conclusion and Future Work}

In this paper, we propose a new video captioning task of describing the off-site audiences or visitors crowd, called crowd video captioning (\textbf{CVC}). In our encoder-decoder system for this task, we use a deep convolutional neural network to extract the features of crowd video and feed them to a language model for crowd description generating. We create a dataset based on WorldExpo'10. On this dataset, our experimental trials prove that our \textbf{CVC} system works well to accomplish this task, and they have achieved high accuracy. It shows that the language model can deeply comprehend the information about the crowd which feature extractors don't understand. 

In our approach, S2VT with features extracted from Inception V3 works better than other methods in the \textbf{CVC} task because our dataset is small and the captions are simple. For future work, these models should be adjusted to fit this fine-grained captioning task for the crowd. The number of videos and the complexity of captions needs to be increased as well in the dataset.

%


\end{document}